\documentclass[letterpaper, 10 pt, conference]{ieeeconf}  

\IEEEoverridecommandlockouts                              

\usepackage{graphicx}
\usepackage{amsmath} 
\usepackage{amssymb}  
\usepackage{duckuments}
\usepackage{booktabs}
\usepackage{cite}
\usepackage{afterpage}
\usepackage{microtype}
\usepackage{url}

\title{\LARGE \bf
Implicit Object Reconstruction With Noisy Data
}

\author{Jad Abou-Chakra$^{1}$, Feras Dayoub$^{2}$, and Niko S\"{u}nderhauf$^{1}$
\thanks{$^{1}$QUT Centre for Robotics, Queensland University of Technology, QLD 4000, Australia. Email: {\tt\small {jad.chakra@hdr.qut.edu.au, niko.suenderhauf@qut.edu.au}}. $^{2}$School of Computer Science, University of Adelaide, SA 5005, Australia. Email: {\tt\small {feras.dayoub@adelaide.edu.au}}. The authors acknowledge the ongoing support of the QUT Centre for Robotics.}
}

\begin{document}

\maketitle
\thispagestyle{empty}
\pagestyle{empty}

\begin{abstract}
Modelling individual objects in a scene as Neural Radiance Fields (NeRFs) provides an alternative geometric scene representation that may benefit downstream robotics tasks such as scene understanding and object manipulation. However, we identify three challenges to using real-world training data collected by a robot to train a NeRF: (i) The camera trajectories are constrained, and full visual coverage is not guaranteed -- especially when obstructions to the objects of interest are present; (ii) the poses associated with the images are noisy due to odometry or localization noise; (iii) the objects are not easily isolated from the background. This paper evaluates the extent to which above factors degrade the quality of the learnt implicit object representation. We introduce a pipeline that decomposes a scene into multiple individual object-NeRFs, using noisy object instance masks and bounding boxes, and evaluate the sensitivity of this pipeline with respect to noisy poses, instance masks, and the number of training images. We uncover that the sensitivity to noisy instance masks can be partially alleviated with depth supervision and quantify the importance of including the camera extrinsics in the NeRF optimisation process.

\end{abstract}



\section{INTRODUCTION}

An appropriate representation of the physical state of a scene can facilitate control and planning functionality that is critical to increasing the utility and flexibility of robotic systems. For example, each of~\cite{jang2018grasp2vec, gao2021kpam, driess2022learning} propose different object representations aiming towards efficient and generalizable manipulation algorithms. Neural Radiance Fields (NeRFs)~\cite{mildenhall2020nerf} are another prospective object representation that densely and compactly capture geometric information from visual data. This representation is learnt from posed images and produces photo-realistic renders from novel viewpoints. For a representation to be used in robotics, it should be constructed reliably from the available sensory input.

\begin{figure}[t]
\centering
\includegraphics[width=\linewidth]{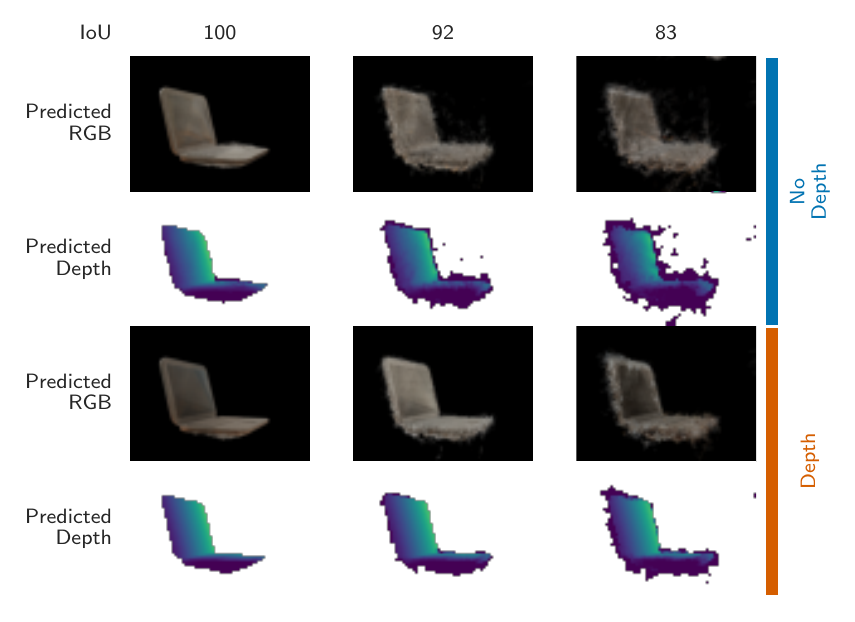}
\caption{A qualitative evaluation for our object-NeRF shows the sensitivity to instance mask noise. The silhouette of the object from a particular viewpoint deviates from its groundtruth with increasing noise levels indicating malformed geometry. Depth supervision alleviates the errors to some degree. }
\label{fig:InstanceExperimentVisuals}
\end{figure}

NeRFs emerged from the computer graphics literature and are usually evaluated based on a photometric reconstruction error. In robotics, the \emph{geometric} reconstruction error is more relevant as downstream tasks -- such as manipulation -- are mostly dependent on shape and not colors. While these two metrics are correlated, it is prudent to have an analysis of NeRFs that considers the influence of errors on the accuracy of the reconstructed shape. 

We aim to give roboticists answers to four practical questions: (i) What shape reconstruction quality can be expected from using an object NeRF? (ii) How accurate do the camera poses need to be and is that achievable with current positioning methods such as visual inertial odometry~\cite{orbslam}? (iii) To what extent does depth information increase the reconstruction quality? (iv) How is the reconstruction quality affected when noisy instance masks are used to isolate objects of interest from their background?

Part of our analysis has some overlap with~\cite{wang2021nerf, lin2021barf} in relation to unknown camera parameters. However, unlike these works, we use a very different NeRF formulation~\cite{mueller2022instant} whose speed in training and evaluation makes it more appropriate for robotics. The difference in the NeRF structure, the use of a geometric metric rather than a photometric one, the object-centric approach, and the robotic lens with which our work is conducted sheds a complimentary light on dealing with inaccurate camera parameters. Furthermore, though there are works that readily use segmentation masks to isolate regions of interest~\cite{yen2022nerfsupervision, driess2022learning2, Tseng2022CLANeRFCA}, there has yet to be an analysis that studies the sensitivity of the reconstruction quality to the fidelity of the masks. It is unclear how a NeRF would react to training data that is partially inconsistent in 3D (as instance masks tend to be). 

Modelling objects as NeRFs requires a strategy of isolating them and of suppressing the inclusion of geometry that does not belong to them. In this work, we implement a pipeline that uses instance masks and object bounding boxes to decompose a scene into multiple object NeRFs. Only loose bounding boxes are needed for the pipeline with the sole requirement that they completely enclose their associated objects. Such bounding boxes can automatically be generated either through the use of an object-based SLAM system -- as in~\cite{quadricslam, cubeslam} -- or by using the available depth data in concert with the instance masks. We then analyze the reconstruction quality the pipeline is able to achieve under various noise levels in the camera poses and the instance masks (Fig.~\ref{fig:InstanceExperimentVisuals}).

\begin{figure}[t]
\centering
\includegraphics[width=0.7\linewidth]{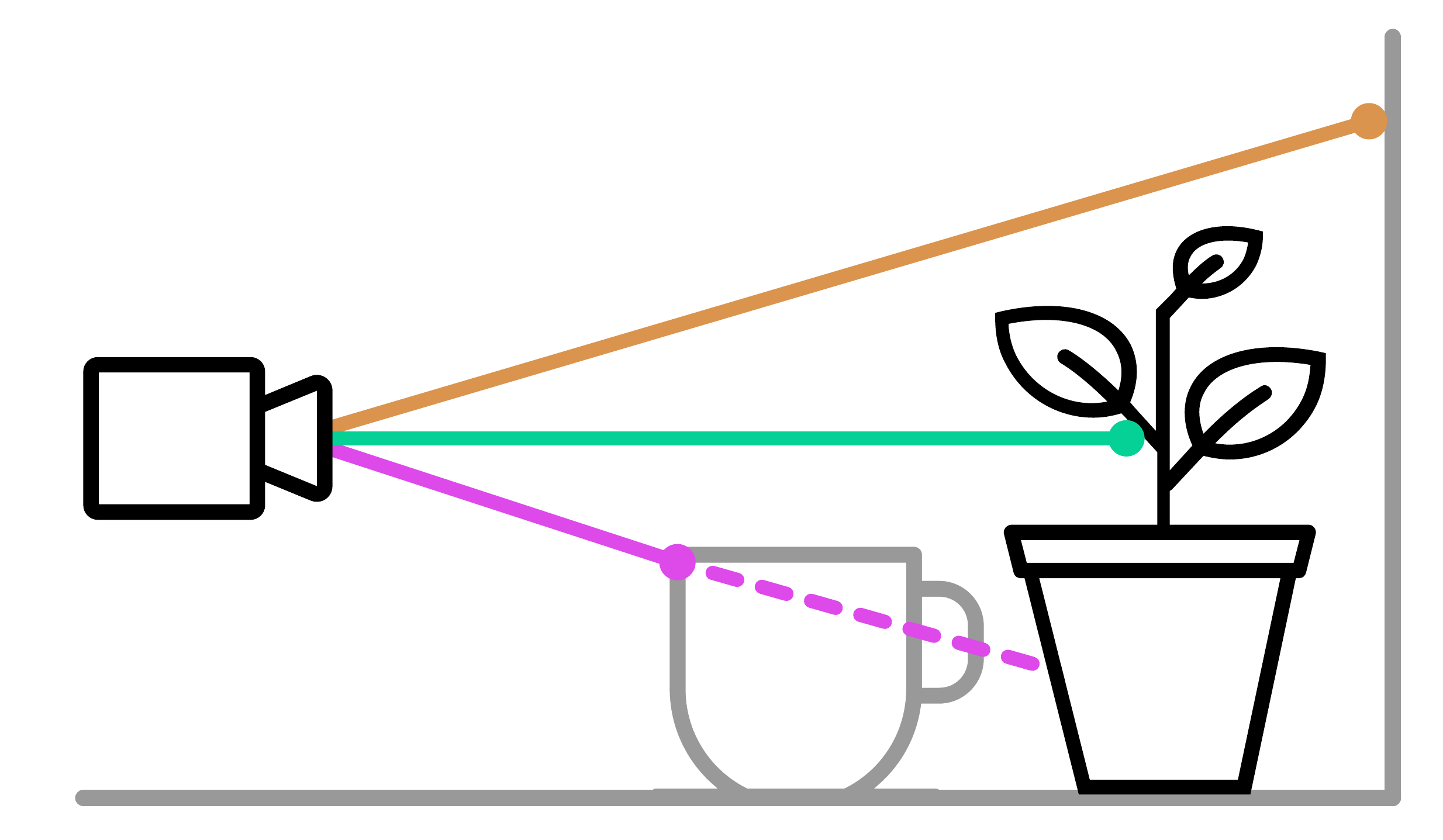}
\caption{
Each training image can be decomposed into three regions through which rays are cast: (i) Rays cast through the negative region -- shown as orange -- represent the background and are optimized towards a zero density distribution. (ii) Rays cast through the positive region -- shown as green -- represent the object and have their computed color optimized towards the ground-truth color in the image.  (iii) Rays cast through the masked region -- shown in pink -- represent possible obstructions and are not included in the training.  }

\label{fig:image_regions}
\end{figure}

\section{RELATED WORK}

Implicit representations~\cite{mildenhall2020nerf, park2019deepsdf, mescheder2019occupancy} compactly capture geometry in continuous functions. Neural radiance fields are examples of these. Mildenhall et al.~\cite{mildenhall2020nerf} show that training a multi-layered perceptron (MLP) to represent the radiance field of a scene allows rendering photo-realistic images from novel views. There have since been many formulations of NeRF~\cite{10.1111:cgf.14505} and only a handful~\cite{mueller2022instant, Chen2022ECCV, yu_and_fridovichkeil2021plenoxels} are fast enough in both training and inference for use in robotics. The difference in speed is mainly due to the method used to represent and encode the inputs. 

Instant-NGP~\cite{mueller2022instant} is currently the fastest formulation of a NeRF and has a well-optimized implementation that is openly available. Therefore, we center our analysis around it. This paper focuses on mapping the sensitivity of the NeRF training process to the number of frames used, the errors in the image poses, and the noise in the instance masks. The aim is to give a non-fragmented analysis of the effect of noise on the training of object NeRFs. While there are several works~\cite{lin2021barf, yen2020inerf, wang2021nerfmm, SCNeRF2021} that specifically tackle the problem of training a NeRF in the presence of camera pose errors, their analysis is conducted on -- and in some cases~\cite{lin2021barf} coupled to -- the Fourier positional encoding that was originally used in~\cite{mildenhall2020nerf}. For example~\cite{lin2021barf} progressively mask the output of the Fourier encoding during training to achieve their results.~\cite{yen2020inerf} require that a NeRF has already been trained.~\cite{wang2021nerfmm} limits their analysis to forward facing scenes. An analysis has therefore not been done on NeRF formulations that use the very different parametric encodings~\cite{mueller2022instant} (which are credited for making NeRFs more suitable to realtime applications). The encodings represent a large part of the NeRF architecture and there is no reason to believe that an analysis on one translates to another. 

Some works~\cite{Zhu2022CVPR, Sucar:etal:ICCV2021} integrate NeRFs into a SLAM system and perform tracking fully within the NeRF optimization process. Our intent in this paper is not to supplant the camera tracking process as other commonly available tracking solutions, such as visual inertial odometry~\cite{leutenegger2015keyframe, orbslam}, are complimentary to the NeRF training process. Our aim is to map how accurate such an auxiliary tracking system needs to be before the NeRF can reliably represent the geometry in its training data. Specifically~\cite{Zhu2022CVPR} report camera tracking accuracy but do not offer a sensitivity analysis as that is not the focus of their work.

Lastly, in this paper, we are interested in creating object-level representations for the possible use in downstream robotic tasks. The simplest way to isolate objects of interest from their surrounding is by using commonly available instance mask networks such as MaskRCNN~\cite{maskrcnn}. Introducing instance masks into the training procedure has been done by several works~\cite{yen2022nerfsupervision, driess2022learning2, Tseng2022CLANeRFCA}, however none have studied how well-formed they have to be. Instance masks are noisy and their use in the training process creates conflicts which the volumetric rending procedure within NeRFs are not designed to handle.

\begin{figure}[t]
\centering
\includegraphics[width=0.9\linewidth, trim={0 0 0 10px},clip]{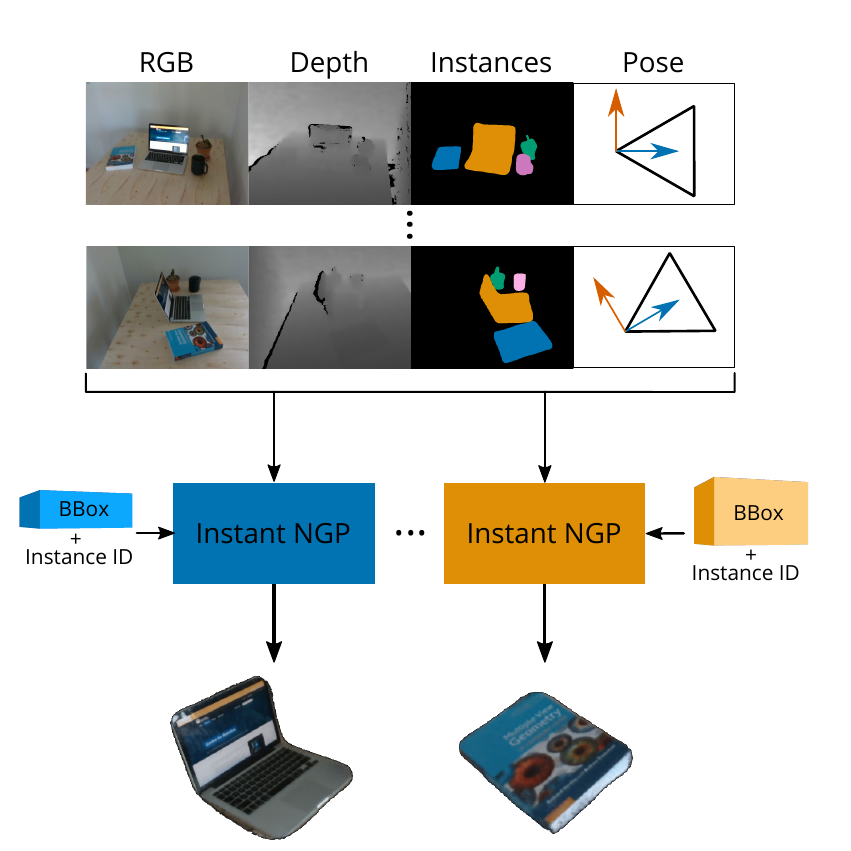}
\caption{
We generate a neural radiance field (NeRF) with a hash encoding (Instant NGP) for each object in the scene. We use noisy instance masks and loose bounding boxes to bound each NeRF and isolate the object from its background. A scene with 4 objects can thus be decomposed into 4 NeRFs, each representing the geometry of a single object.
}
\label{fig:architecture}
\end{figure}

\section{PRELIMINARY}
A NeRF is a continuous representation of a 3D scene that maps a point $\boldsymbol{x}_i \in \mathbb{R}^3$ and a unit-norm view direction $\boldsymbol{d}_i \in \mathbb{R}^3$ to a colour $\boldsymbol{c}_i \in \mathbb{R}^3$ and a density value $\sigma_i \in \mathbb{R}$. 
NeRFs are trained by casting rays from a camera center through the image plane and progressively sampling points along them~\cite{mildenhall2020nerf}. A ray is parameterized as $\boldsymbol{r}(t) = \boldsymbol{o} + t\boldsymbol{d}$ where $\boldsymbol{o} \in \mathbb{R}^3$ is the ray origin and $\boldsymbol{d} \in \mathbb{R}^3$ is the ray direction. Sampling $N+1$ points along the ray and defining $\delta_i = t_{i+1} - t_i$ for $i\in\{0,...,N\}$, the expected ray colour $\hat{C}(\boldsymbol{r})$ and the expected distance a ray will travel $\hat{D}(\boldsymbol{r})$ are given by:

\begin{gather}
    \hat{C}(\boldsymbol{r}) = \sum_{i=1}^{N}w_i\boldsymbol{c}_i  \quad\text{and}\quad 
    \hat{D}(\boldsymbol{r})=\sum_{i=1}^{N}{w_i t_i} \nonumber \\ 
    \text{where: }\\ w_i = T_i(1-\exp(-\sigma_i\delta_i)) \text{ and } T_i = \exp{\left(-\sum_{j=1}^{i-1}{\sigma_j\delta_j}\right)}
    \label{eq:render}
\end{gather}

Refer to~\cite{tagliasacchi2022volume, 468400} for an in-depth derivation.

\section{METHOD}
Given a set of images $I_i$ and their corresponding poses $X_i$, depth maps $D_i$, and instance masks $S_i$, we aim to construct a separate NeRF for every object $j$ present in the scene -- see Figure~\ref{fig:architecture}. A ray $\boldsymbol{r}$ that intersects an image plane $i$ at a pixel coordinate $(u, v)$ is associated with the color $\boldsymbol{c}_{gt}(\boldsymbol{r}) = I_i(u,v)$, the depth $d_{gt}(\boldsymbol{r}) = D_i(u,v)$, and the instance ID $S_i(u,v)$.

To construct a single-object NeRF, it is necessary to suppress geometry that does not belong to the object -- such as the background and other objects. To this end, rays that are known to hit the object should encourage geometry formation along them. Conversely, rays that do not hit the object should discourage it. Rays that are obstructed from hitting the object by another are not included in the training because their ground-truth values are not known -- the colors that correspond to them are not of the object of interest but rather of the object that is obstructing it. Therefore, rays are cast from each training image through three different regions (illustrated in Figure \ref{fig:image_regions}): (i) the negative region -- corresponding to the background -- that discourages geometry formation, (ii) the positive region -- corresponding to the object -- that promotes it, and (iii) the masked region -- corresponding to potential obstacles -- that does neither. 

The rays $\mathcal{R}_{p}$ cast through positive regions are given ground-truth targets from the training images as is done in conventional NeRF training. They are identified as those rays that have a corresponding instance ID equal to the object's ID. The rays $\mathcal{R}_{m}$ passing through the masked region have no influence on training. These are the rays whose associated instance ID belong to an object other than the one being mapped and intersect that object's bounding box before intersecting the NeRF's. The remaining rays $\mathcal{R}_{n}$ are the ones cast through the negative regions. If every point on one such ray has a density of 0, the rendered color would be $\boldsymbol{0}$ (black). The inverse, however, is not true. Therefore, to avoid the edge case of creating black geometry instead of no geometry, the rays are given a constantly changing colour target to encourage the densities towards zero and the colors away from it (as is done in~\cite{mueller2022instant}). This method assumes that the geometries obstructing the training views are recognized by the instance masks. 

Our final photometric loss is formulated as:

\begin{equation}
    L_{\text{rgb}} = \sum_{\boldsymbol{r} \in \mathcal{R}}\| \boldsymbol{e}_{\text{rgb}}(\boldsymbol{r}) \|_{2} 
\end{equation}
where
\begin{equation}
    \boldsymbol{e}_{\text{rgb}}(\boldsymbol{r}) = 
    \begin{cases}
        \hat{C}(\boldsymbol{r}) - \boldsymbol{c}_{\text{gt}}(\boldsymbol{r}) \quad &\text{if} \quad \boldsymbol{r} \in \mathcal{R}_p \\ 
        \hat{C}(\boldsymbol{r}) - \boldsymbol{c}_{\text{random}}  \quad &\text{if} \quad \boldsymbol{r} \in \mathcal{R}_{n} \\ 
        0 \quad &\text{if} \quad \boldsymbol{r} \in \mathcal{R}_{m}
    \end{cases}
\end{equation}

$\mathcal{R}$ is the set of all rays that pass through the training images. $\boldsymbol{c}_{gt}(\boldsymbol{r})$ is the color of the pixel in the training image that the ray $\boldsymbol{r}$ intersects. $\boldsymbol{c}_{\text{random}}$ is a vector drawn from a uniform distribution $U(0, 1)$. 

In some experiments, we show how depth supervision
affects the results. In those cases, the depth loss is given as:

\begin{equation}
    L_{\text{depth}} = \sum_{\boldsymbol{r} \in \mathcal{R}_d}| \boldsymbol{e}_{\text{depth}} |
    \label{eq:depth_loss}
\end{equation}
where
\begin{equation}
    \boldsymbol{e}_{\text{depth}}(\boldsymbol{r}) = 
    \begin{cases}
        \hat{D}(\boldsymbol{r}) -- \boldsymbol{d}_{\text{gt}}(\boldsymbol{r}) &\text{if } \boldsymbol{r} \in \mathcal{R}_p \text{, } T_N < 10^{-4},  \\ 
        & \text{and } \boldsymbol{d}_{\text{gt}}(\boldsymbol{r}) > 0 \\
        0 &\text{otherwise}
    \end{cases}
\end{equation}

$\mathcal{R}_d$ is the set of rays for which ground-truth depth information is available. The method is not sensitive to the empirically chosen threshold condition $T_N<1e^{-4}$.

\textbf{Joint Optimization:} The learnable parameters $\Phi$ of the NeRF framework and the poses $X$ of the cameras are jointly optimized. 
\begin{equation}
    \Phi^*, X^* = \arg \min_{\Phi, X} L_{\text{rgb}} + w_{\text{depth}}L_{\text{depth}}
\end{equation}

\begin{table}[b]
\caption{Instant-NGP network configuration}
\centering
\begin{tabular}{@{}lr@{}}
\toprule
Parameter                              & Value   \\
\midrule
Number of levels                       & $16$      \\
Hash table size                        &  $2^{19}$       \\
Number of feature dimensions per entry & $2$       \\
Coarsest resolution                    & $16$      \\
Finest resolution                      & $2048$        \\
\bottomrule
\end{tabular}
\label{tab:config}
\end{table}

\textbf{Implementation Details:} Instant-NGP~\cite{mueller2022instant} with a multi-resolution hash encoding is used as the underlying NeRF model because it trains in close to real-time making it the most suitable for robotic applications. The implementation provided by~\cite{mueller2022instant} is extended to add support for a depth loss and to isolate objects based on an instance ID found in the masks. For details on Instant-NGP itself and its hash encoding, the reader is referred to~\cite{mueller2022instant}. The Instant-NGP configuration used is described in Table \ref{tab:config}. The authors of~\cite{mueller2022instant} optimize each training pose with Adam~\cite{kingma2014adam} configured with $\beta_1 = 0.9$, $\beta_2 = 0.99$, $\epsilon = 10^{-9}$. The learning rate is exponentially decayed from  $3.3 \cdot 10^{-4}$ to $10^{-5}$. All models are trained for 2000 steps. $w_{\text{depth}} = 3.0$ when the depth loss is included.

\section{EVALUATION}
\textbf{Metrics:} In contrast to those works that use NeRFs for novel view synthesis, we evaluate primarily on the accuracy of the geometry rather than that of the color. We differentiate between areas which have been correctly categorized as part of the object of interest and areas which have not. For correctly categorized areas, we use the mean average depth error (MAE) between the ground-truth and rendered depth maps to measure accuracy. We also use an intersection-over-union (IoU) between the ideal and rendered instance masks to measure how much of the total shape is represented in the NeRF.

\begin{figure}[t]
\centering
\includegraphics[width=\linewidth, trim={0 5px 0 20px}, clip]{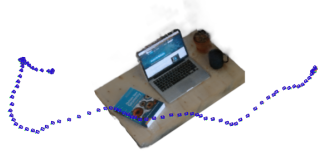}
\caption{A real scene -- from our dataset -- with four objects of interest is explored by a camera moving along a constrained trajectory  -- shown in blue. The path is calculated by ORBSLAM~\cite{orbslam} as is typical for many robotic applications. Object NeRFs extracted from this scene must deal with the limited viewpoints afforded by the trajectory, the objects obstructing each other, and noise in the computed poses and instance masks.}

\label{fig:RealScene}
\end{figure}

\begin{figure}[t]
\centering
\includegraphics[width=\linewidth]{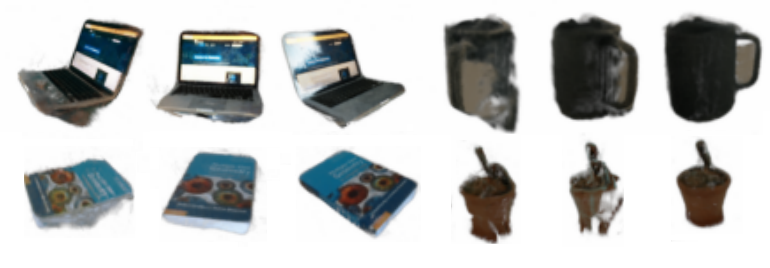}
\caption{Renders from four object NeRFs extracted from a real scene show various geometry artifacts that may be due to insufficient view coverage, noisy instance masks, and inaccuracies in the poses. The training data is taken from a constrained trajectory and instance masks are produced by MaskRCNN. Depth supervision is not used in this example. Large simple geometries -- as seen in the laptop and book renders -- are reconstructed more accurately than their counterparts. In general, the objects are well-isolated from their background, however attributing the artifacts to different sources of error is difficult in real scenes. This begs the question: ``To what extent does each noise factor contribute to imperfections in the NeRF?" We answer that question in our analysis. }

\label{fig:RealSceneVisuals}
\end{figure}

The method is run on the real scene shown in Figure \ref{fig:RealScene} and qualitative data is collected and shown in Figure \ref{fig:RealSceneVisuals}. A camera is used to capture 100 images along a constrained trajectory. Loose bounding boxes are manually estimated around objects in the scene. Bounding boxes can easily be obtained from object-based SLAM systems~\cite{quadricslam, cubeslam} or from 3D object detection networks~\cite{ahmadyan2021objectron}, however we choose not to complicate the analysis further by involving additional systems -- especially because the quality of the bounding box does not affect the results (The only requirement is that the bounding box fully enclose the object -- the quality of the fit is irrelevant). Camera positions are initialized with ORBSLAM3~\cite{orbslam}. Lastly, instance masks are given by MaskRCNN~\cite{maskrcnn}. 

\begin{figure}[t]
\centering
\includegraphics[width=\linewidth,height=3cm, trim={0 20px 0 0},clip]{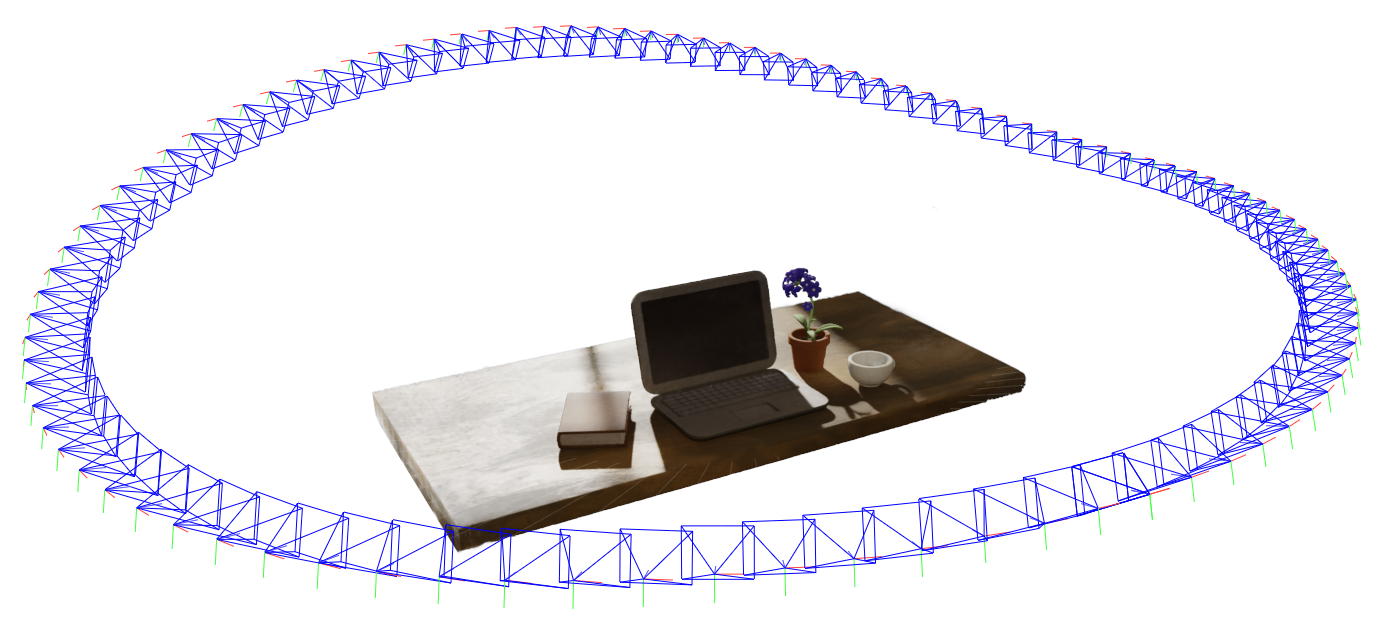}
\caption{The synthetic scene from our dataset that mimics the setup of the real scene.}
\label{fig:ConstrainedImage}
\end{figure}

Because ground-truth data is unavailable when using real-world data, it is difficult to quantitatively assess the method. Therefore, we replicate the real scene in Blender~\cite{blender} -- Figure \ref{fig:ConstrainedImage} -- and measure the reconstruction quality of the object NeRFs under various conditions. The affects of using depth supervision, ORBSLAM, and MaskRCNN on the synthetic scene with a constrained camera trajectory are shown in Table \ref{tab:ConstrainedExperiment}. We observe that most of the error is due to the use of MaskRCNN. Use of non-ideal instance masks results in a doubling of the reconstruction error and a 7 to 12 percent drop in the IoU metric. This indicates that noisy instance masks have a severe effect on the quality of the representation.

\begin{table}[t]
\caption{Effect of noisy data and depth supervision on the reconstruction of objects from a constrained trajectory.}
\begin{tabular}{lllrr}
\toprule
Masks & Depth & Poses & Depth MAE (cm) & IoU (\%) \\
\midrule
 Ideal & True & Ideal  &          $0.5\pm0.3$ &  $98\pm1.5$ \\
       & True  & ORB-SLAM &     $0.6\pm0.3$ &  $98\pm0.6$ \\
       & False & Ideal &        $1.1\pm0.3$ &  $98\pm0.8$ \\
       & False & ORB-SLAM &     $1.4\pm0.3$ &  $98\pm1.1$ \\
MaskRCNN & True  & ORB-SLAM &     $1.2\pm0.1$ &  $85\pm5.8$ \\
         &       & Ideal &        $1.3\pm0.2$ &  $85\pm7.4$ \\
         & False & ORB-SLAM &     $2.3\pm0.4$ &  $83\pm6.3$ \\
         &       & Ideal &        $2.3\pm0.4$ &  $82\pm7.8$ \\
         
\bottomrule
\end{tabular}
\label{tab:ConstrainedExperiment}
\end{table}

To further analyze the relationship between noise in the instance masks and poses, we construct three experiments on a synthetic dataset. Section \ref{sec:Baseline} begins by determining an upper bound on reconstruction quality in ideal conditions. Section \ref{sec:Instance Noise} studies the influence of noisy instance masks. Section \ref{sec:pose_experiment} analyzes the robustness to increasing inaccuracies in camera poses.

\begin{figure}[t]
\centering
\includegraphics[width=0.9\linewidth]{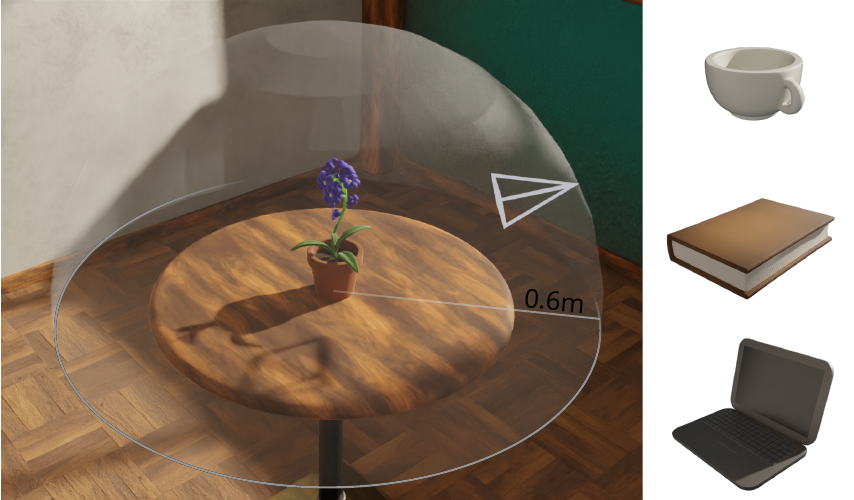}
\caption{The Blender Cube Diorama scene that is used in the baseline experiment, the instance noise experiment, and the pose error experiment. Training images are generated from a camera looking at the center of the table and placed on the sphere shown. The four table-top models ``bluebell", ``laptop", ``book", and ``cup"  are used to calculate confidence intervals.}
\label{fig:dataset}
\end{figure}

\textbf{Datasets:} Simulated and real datasets used in this paper are made available at~\cite{AbouChakraNeRFCubeDiorama2022}. Simulated experiments below are derived from the Blender Cube Diorama scene~\cite{blender}. The Blender scene is packaged with a number of assets that make it easy to conduct controlled experiments with groundtruth data and for others to recreate the experiments. We test on four objects ``Laptop", ``Tea Cup", ``Bluebell", and ``Book". These are shown in Figure \ref{fig:dataset}. The particular items are chosen because they can also be segmented by MaskRCNN~\cite{maskrcnn} pre-trained on the MS COCO dataset~\cite{coco}. The models vary in shape and size and present different challenges to the training. The ``Bluebell" model, for example, is composed of many thin structures, whereas the ``Tea Cup" is relatively small and has a hard to learn cavity. All images have a resolution of 640x480 -- matching the real dataset used earlier.

\subsection{Baseline}
\label{sec:Baseline}
Before evaluating the effect of various noise levels on NeRF training, a baseline is established in an ideal scene with perfect data and well-distributed cameras. Four scenes are constructed; each having an object from Figure \ref{fig:dataset} placed on a table. A NeRF is trained using a set of images and instance masks taken from the top half of a sphere looking at the center of the object. We show how increasing the number of views, utilizing a depth loss, and changing the distance between the object and the cameras affects the reconstruction. The average depth error and the IoU are measured across a test set of 50 novel views from the same top hemisphere. 

\begin{figure*}[t]
\centering
\includegraphics[width=0.93\linewidth]{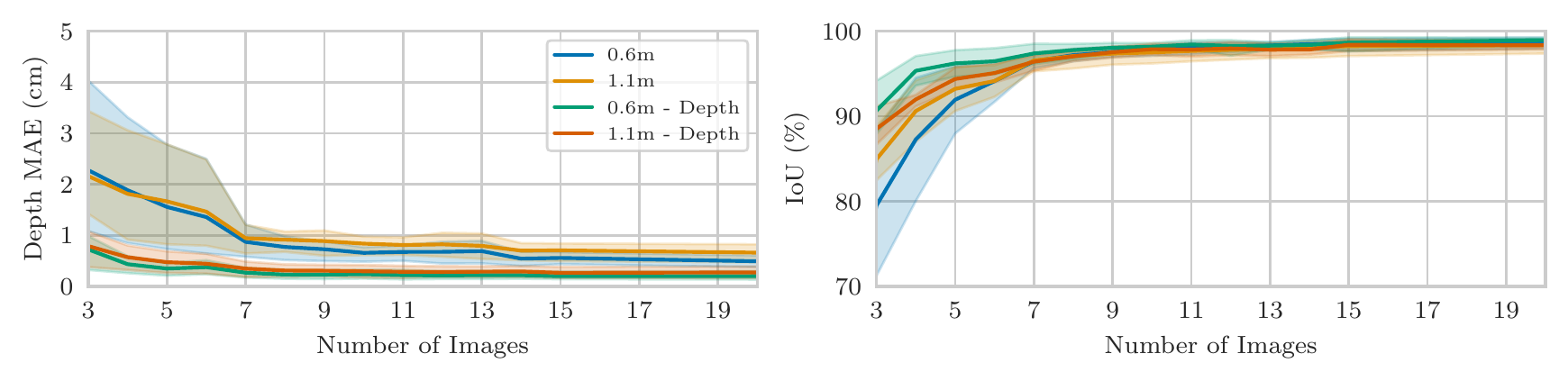}
\caption{NeRFs -- representing one of four tabletop objects placed in an indoor scene -- are trained from an increasing number of well-distributed posed images with groundtruth instance masks. The images are taken from the top half of a sphere with radius 0.6m and 1.1m. Depth supervision is enabled for two of the experiments. The tabletop objects can be reconstructed with an accuracy under $1\,\text{cm}$ at $98\%$ IoU. The number of training images has a diminishing effect on the quality of the recontruction whereas depth supervision gives an overall increase in accuracy ($<0.5\,\text{cm}$ at $98\%$ IoU).
}
\label{fig:BaselineExperiment}
\end{figure*}

\textbf{Results:} Under ideal conditions, Figure \ref{fig:BaselineExperiment} shows a reconstruction accuracy of 0.2~cm to 0.8~cm at 98\% IoU can be achieved. Overall, increasing the number of images past a certain point has diminishing returns. Including a depth loss in the training results in an absolute increase in accuracy. Lastly, the farther away the image planes are from the object, the less accurate the reconstruction.

\subsection{Instance Noise}
\label{sec:Instance Noise}
Instance masks are used to identify objects within an image. In contrast to synthetic scenes, masks output from MaskRCNN or other instance segmentation networks produce noisy output. The noise can be thought of as being applied in 2D. Therefore, masks of the same objects taken from different views are not consistent in the 3D world. In other words, rays that intersect an object at the same point may disagree on whether they should be classified as hitting the object or not. This experiment investigates how robust NeRFs are to this kind of noise. 

\begin{figure}[t]
\centering
\includegraphics[]{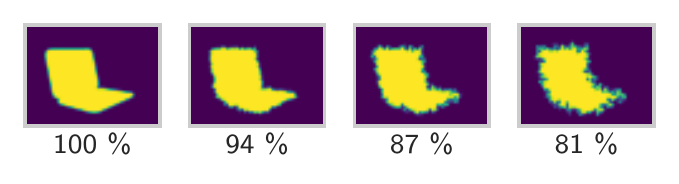}
\caption{Example outputs from the instance noise generator. This is used to controllably introduce noise unto ideal instance masks and analyze its affect on the NeRF reconstruction. The numbers shown are the intersection-over-union of the resultant mask relative to the groundtruth.}
\label{fig:InstanceNoise}
\end{figure}

IoU is used as a measure of noise on the instance mask. Noise is introduced to an ideal mask by iteratively adding circular patches to the edge of the instance boundary until the IoU with the groundtruth is within 1\% of the desired value. Exemplary outputs from this method are shown in Figure \ref{fig:InstanceNoise}. The experiment from Section \ref{sec:Baseline} is repeated with varying degrees of noise across the instance masks. We are careful to remove and add patches to the instance mask with equal probability. This provides an unbiased noise model.

\textbf{Results:} Figure \ref{fig:InstanceExperiment} show that NeRFs are highly sensitive to noise in the instance masks. Increasing the number of images does not give an appreciable robustness to the noise. Adding depth supervision seems to give the NeRF more robustness. In general, however, noise in the instance masks cause a rapid deterioration in the reconstruction quality.

\subsection{Camera Pose Errors}
\label{sec:pose_experiment}

Robotic systems localize cameras with varying degrees of accuracy. We investigate how resilient NeRF training is to the noise. In this experiment, reconstruction quality is evaluated under increasing levels of perturbations to the camera poses. Rotation and translation inaccuracies are studied independently. Translation noise is introduced by adding a vector $\eta \sim N(\boldsymbol{0}, \text{diag}(\sigma_t^{2}, \sigma_t^{2}, \sigma_t^{2}))$ to the camera location. Rotation noise is introduced by twisting a camera pose about an arbitrarily chosen axis with an angle $\theta \sim N(0, \sigma_r^2)$.

Optimizing the camera poses as part of the training can cause a general drift in the poses. For example, if the poses drift upwards by $1\,\text{cm}$ then the reconstruction itself is also shifted by $1\,\text{cm}$ from the original coordinate frame. Consequently, the test set cannot be used for evaluation until the new coordinate frame is registered with the old one. To avoid this complication, we evaluate on the training set. This is an acceptable proxy because geometry formation is predominantly informed by color whereas the metric is derived from depth. The baseline experiment is repeated with pose noise applied.

\begin{figure*}[t]
\centering
\includegraphics[width=0.93\linewidth]{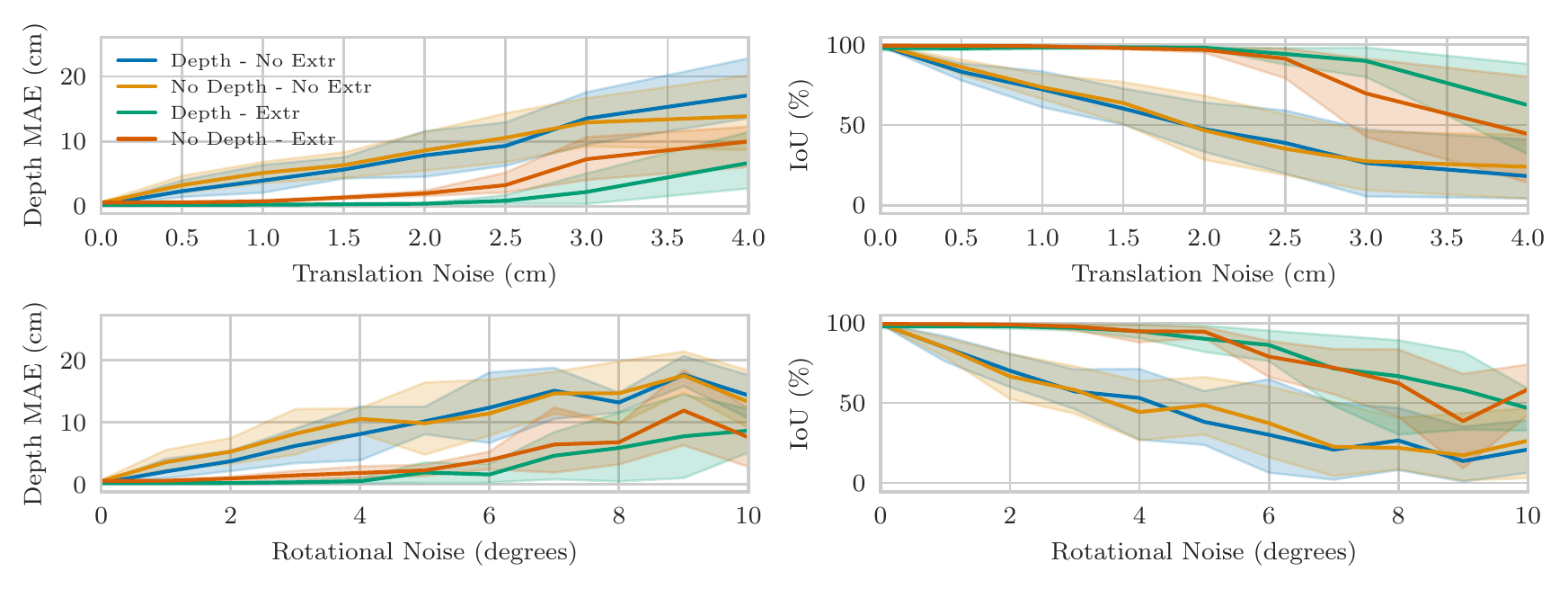}
\caption{Well-distributed images with ideal instance masks and inaccurate poses are used to reconstruct the same tabletop objects from the baseline experiment. Translation and rotation errors are treated separately. The robustness to both is shown when depth supervision is included and when the extrinsics -- labelled ``Extr'' in legend -- are allowed to be optimized.  }
\label{fig:PoseExperiment}
\end{figure*}

\textbf{Results:} Figure \ref{fig:PoseExperiment} shows that training on extrinsics gives a much needed robustness to localization errors. The NeRFs in our example can tolerate up to $2\,\text{cm}$ of translation noise and 3 degrees of rotation error.

\section{DISCUSSION}
\begin{figure*}[t]
\centering
\includegraphics[width=0.93\linewidth]{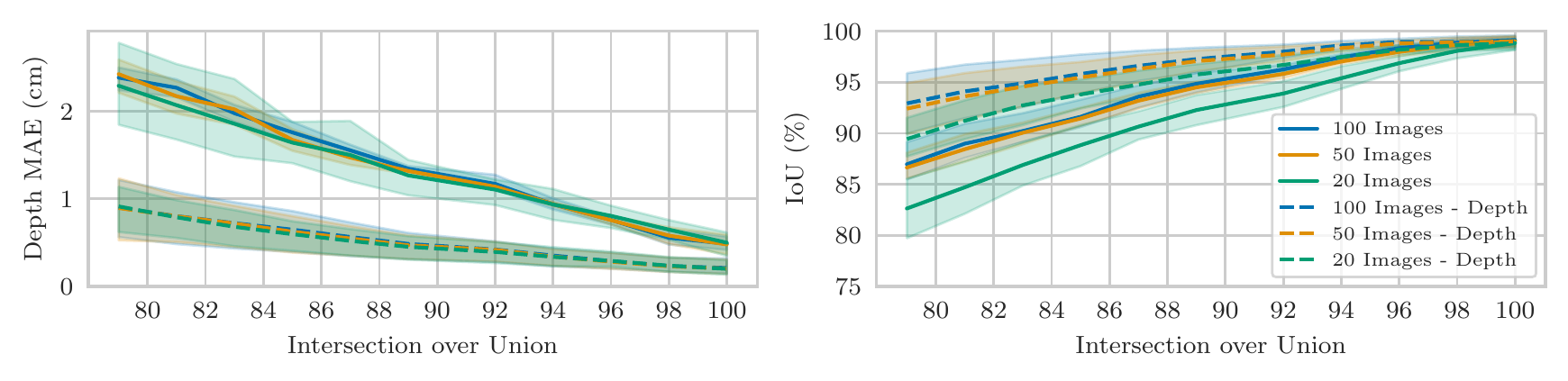}
\caption{Well-distributed images with groundtruth poses and noisy instance masks are used to reconstruct the same tabletop objects from the baseline experiment. The reconstruction quality deteriorates quickly with increasing noise levels. Depth supervision is more effective than increasing the number of training images at slowing the deterioration.}
\label{fig:InstanceExperiment}
\end{figure*}

Baseline experiments show that object NeRFs can provide a dense and accurate geometric representation. For the objects tested, a 4mm accuracy with 100\% IoU can ideally be achieved. The issues of bounding and isolating objects from a scene are partially solved by our method through the use of instance masks and bounding boxes. However, the isolation process is not trivial. Given a set of instance masks that should agree in the 3D world but do not, what strategy should be used to minimise the affect of the inconsistency? 
Results in Figure \ref{fig:InstanceExperiment} do not indicate that the NeRF formulation in its current form can deal with corrupted data reliably. Therefore finding a more robust procedure is an open problem.

\section{CONCLUSIONS}

In this work, we demonstrated and analyzed a pipeline for decomposing a scene into object NeRFs. We found they are capable of millimeter level accuracy when used with ground-truth data. In robotic applications, camera poses are likely to have errors in the centimeter range. For NeRFs to be a usable object representation and have a reliable reconstruction quality, the camera extrinsics should be included in the optimization. We uncovered a sensitivity to noisy instance masks that is to some extent alleviated by the presence of depth data and more training images. This raises a vital question for future research which we believe is important if this representation is ever to be used as an object representation in robotics: How can NeRFs be trained to be robust to data that is not consistent in 3D? This paper provides a non-fragmented analysis on the influence of noise and errors in the training data on the geometric reconstruction quality of a NeRF.  

\newpage
\clearpage
\clearpage




\addtolength{\textheight}{0cm}   
\bibliographystyle{IEEEtran}
\bibliography{references}
\end{document}